# Accelerating Recurrent Neural Network Training using Sequence Bucketing and Multi-GPU Data Parallelization

Viacheslav Khomenko [1], Oleg Shyshkov [1], Olga Radyvonenko [1], Kostiantyn Bokhan [1]

[1] Samsung R&D Institute Ukraine (SRK), 57, L'va Tolstogo Str., Kyiv, 01032, Ukraine, v.khomenko@samsung.com

*Abstract*—An efficient algorithm for recurrent neural network training is presented. The approach increases the training speed for tasks where a length of the input sequence may vary significantly. The proposed approach is based on the optimal batch bucketing by input sequence length and data parallelization on multiple graphical processing units. The baseline training performance without sequence bucketing is compared with the proposed solution for a different number of buckets. An example is given for the online handwriting recognition task using an LSTM recurrent neural network. The evaluation is performed in terms of the wall clock time, number of epochs, and validation loss value.

## I. Introduction

Deep neural networks have recently proven to be successful in pattern recognition tasks. The Recurrent Neural Network (RNN) is a subclass of neural networks defined by presence of feedback connections.

Long Short-Term Memory (LSTM) [1] RNNs perform better on tasks involving long time lags compared to traditional RNNs. The gating mechanism permits LSTM to bridge long time lags between relevant events ($10^3$ time steps and more). Gated Recurrent Unit (GRU) networks [2] have similar ideology to an LSTM, but they speed up training due to architectural simplifications.

The ability of RNNs to memorize historical data makes them a powerful sequence-modeling tool. They have found applications in pattern recognition and classification tasks where inputs and outputs are sequences: online handwriting recognition [3], document analysis [4], sentiment analysis [5], speech recognition [6] and synthesis [7], language modeling [8].

However, RNN training on a big amount of data is still a challenging problem. The aim of the paper is to demonstrate an effective approach to accelerate RNN mini-batch training on big amount of data.

This paper is organized as follows. The related works overview is given in section 2. Section 3 describes the training algorithm. The experimental evaluation is given in section 4. Then, the results are discussed, and the evaluation of the proposed sequence bucketing algorithm against the conventional sequence shuffling is presented.

## II. Background and related work

The problem of accelerating the RNN mini-batch gradient descent training is widely discussed in the literature in the last years [8, 9]. Some works consider adaptive learning algorithms with heuristics for tuning of learning parameters (learning rate, weight decay, etc.) to improve convergence of model training [8].

Many researchers have focused their efforts on experiments with different network architectures and parallelization techniques [6, 9 and 10].

Training parallelization and a two-stage network structure for RNN [9] allow to speed-up training. However, the two-stage architecture gives substantial acceleration mainly when the number of outputs of the network is sufficiently large ($10^3$ or more).

The BlackOut [11] approach allows to accelerate training for even larger vocabularies ($10^6$ outputs). It relies on weighted sampling strategy, employs a discriminative training loss and is applied only to the *softmax* output layer, in contrast to DropOut [12], which is typically applied to the input and hidden layers. The application of BlackOut is also limited to networks with large output size.

In its turn, the curriculum learning [13] consists in organizing training samples in a meaningful way rather than in purely random order. It improves LSTM training on the program evaluation and memorization tasks [14].

It is commonly known that labeling of unsegmented sequence input data is a ubiquitous problem in the real-world sequence learning. It is particularly common in perceptual tasks (e.g. handwriting, speech or gesture recognition), where noisy real-valued input sequences are annotated with non-aligned strings [15]. Since Connectionist Temporal Classification (CTC) networks gained use in RNN training as a sequence alignment technique, the problem of RNN training using input sequences of different length turned out to be more important and affecting training speed.

Usually, sequences are grouped into mini-batches. The length of the longest sequence in the batch thus defines the computational complexity of the training. Most of benchmark datasets for perceptual machine learning tasks (TIMIT [16], UNIPEN [17], IAMonDo [18]) contain recordings of different length. Batch grouping algorithms could be useful for organizing training data [19, 20]. However, the following two problems arise in this case:

1. Finding the optimal batch clustering by sequence lengths.

2. Balancing between input data streamlining and the need of shuffling training data sequences before RNN training.

In the next sections, we present and evaluate the RNN training approach with an effective sequence bucketing that solves problems mentioned above.

## III. TRAINING ALGORITHM BASED ON SEQUENCE BUCKETING AND MULTI-GPU DATA PARALLELIZATION

We propose the RNN model training algorithm that runs in parallel on multiple Graphical Processing Units (GPUs). The developed solution uses a map-reduce approach for parallel computing of individual models by sub-partitioning training data. The training data is shuffled before every epoch and is equally redistributed between different GPU processes. Each training process applies batch bucketing optimization scheme by clustering training sequences considering input lengths. Final model parameters are obtained by reducing results of each training process.

The proposed training workflow is presented in the diagram (Fig. 1).

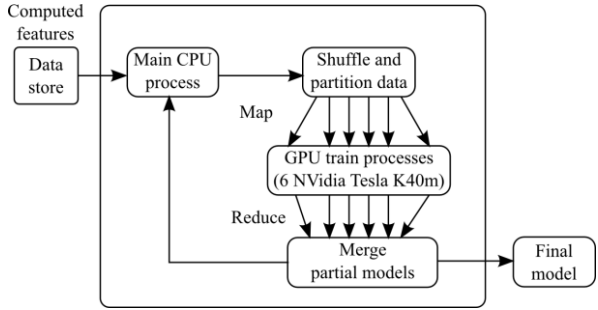

Figure 1. Proposed training algorithm with multi-GPU data parallelization. The sequence bucketing occurs in each GPU process

### A. Sequence bucketing algorithm

We accelerate the training on the individual GPU by sequence bucketing that deals with the problem of large variation of input sequence lengths. The empirical distribution of input sequence lengths and an example of clusterization for the number of buckets $Q = 6$ are shown in Fig. 2.

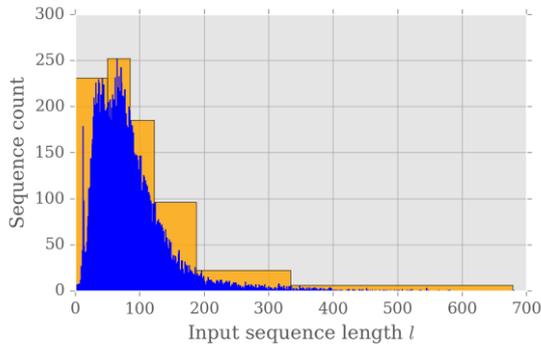

Figure 2. Lengths of input sequences for the online handwriting recognition task and bucketing for $Q = 6$: blue is the distribution of the input sequence lengths; orange are optimal buckets

The bucketing can be described as an optimization problem. Let $S = \{s_1, s_2, ..., s_n\}$ be the set of sequences and $l_i = |s_i|$ is the length of sequence $i$. Each GPU processes sequences in a mini-batch in a synchronized parallel manner, so processing time of a mini-batch $B = \{s_1, s_2, ..., s_k\}$ is proportional to $O(\max_{i \in 1,...,k} l_i)$ and processing time of whole set is expressed as:

$$T(S) = O(n/k \cdot \max_{i \in 1,...,n} l_i) \quad (1)$$

The minimum and maximum sequence length in a mini-batch might be very different if sequences were shuffled randomly before splitting. As a result, a GPU does additional work by processing empty tails of shorter sequences. To overcome this flaw and decrease processing time, we recommend an algorithm that optimizes batch clustering.

Let's call bucketing a process when we cluster all sequences into $Q$ buckets by their lengths, where $Q$ is some small positive integer number. Let $S_i = \{s_{j_i}, ..., s_{j_{i+1}-1}\}$ be the $i$ th bucket. For every bucket, we perform mini-batch training. The processing time of the whole set will become:

$$T(S) = O\left(\sum_{i=1}^{Q} T(S_i)\right) = O\left(n/k \cdot \sum_{i=1}^{Q} \max_{p \in j_i \cdots j_{i+1}-1} l_p \Big/ Q\right) \quad (2)$$

The dynamic programming algorithm is used to find optimal bucket sizes. The bucket sizes only depend on sequence lengths, so we use an array $f$ that stores a number of sequences for each length.

We use the following notations:
- $Q$ is the desired number of buckets;
- $f[l]$ is the number of frequencies with input length $l$;
- $dp[i][k]$ is the best score of bucketing if first $i$ elements were cut into $k$ groups;
- $dp[0][0]$ is set to 0;
- $dp[i][0]$ is set to $INF$;
- $dp[i][k] = \min_{j=1}^{j<i-1}\left(dp[j][k-1] + i \cdot \sum_{t=i+1}^{i} f[t]\right)$;
- $prevDp[i][j]$ is the end index of the $i - 1$ bucket when first $j$ elements were split into $i$ buckets.

The pseudo-code of the proposed algorithm for sequence bucketing is presented below:

```
procedure DYNAMICBUCKETING (Q, f)
    n ← length(f)
    for q = 1 to Q do
        for i = 1 to n do
            curSum ← f[i]
            for j = i - 1 downto 0 do
                val ← curSum · i + dp[q - 1][j]
                if val < dp[q][i] then
                    dp[q][i] ← val
                    prevDp[q][i] ← j
                end if
                curSum ← curSum + f[j]
            end for
        end for
    end for
    curId ← n - 1
    bests ← []
    for i = Q downto 1 do
        bests.push_front(curId)
        curId ← prevDp[i][curId]
    end for
    return bests
end procedure
```

The algorithm optimization result for the given distribution of input sequence lengths and desired number of buckets is presented in Fig. 2.

## B. Parallel training of recurrent neural networks

For the parallel training of the RNNs on GPU we propose the following algorithm which allows massive parallel model training and can scale up to a large number of GPUs:

1. Initialize base model parameters.
2. Build $Q$ models (for optimal input sequence lengths) and serialize models to storage, for example, the file system. Parameters of each model are initialized randomly.
3. Generate training data (1st epoch) or re-generate training data by re-shuffling between portions of data for following epochs. The number of data portions is equal to the number of GPUs. This is possible under the assumption that the amount of training data is sufficient [9].
4. For each of the training data portions, spawn individual training processes. The process iterates over the mini-batches in the data portion. The batches are formed considering input sequence lengths. The appropriate models are selected. The parameter update rule of the individual model is ADADELTA [21].
5. Each of the training processes returns model parameters. The aggregation of parameters is done in the main process according to the model update rule proposed in [9]. We have found that setting parameter $\alpha = 1$ gives the best results for our model and leads to the following equation:

$$V(t) = V_\alpha(t) - \beta \cdot V(t-1) \quad (3)$$

where $V(t)$ are the parameter values at the current epoch; $V(t-1)$ are the parameter values at the previous epoch (or initial values for the first epoch); $V_\alpha(t)$ are the mean parameter values over parallel models; $\beta = 1 \cdot 10^{-6}$ is the regularization term that leads to weight decay proportionally to their values.

## IV. EMPIRICAL EVALUATION

### A. Experimental setup

The system was evaluated on online handwriting recognition task. The raw data contain 1 Gb samples of Afrikaans and English languages in binary form.

The dataset was collected on Samsung Galaxy S-Note devices with stylus input. The validation set is created from 5% of randomly selected samples of different length. The dataset contains textual labels that serve as a reference to output sequences. However, these labels are not explicitly aligned with input handwriting stroke sequences. At every epoch, the system was first fed with shorter sequences (first bucket), and then gradually the bucket number increased.

The LSTM model was trained with CTC cost function using Theano [22] and Lasagne [23] frameworks.

The recurrent neural network model training procedure was evaluated on a rack of 6 NVidia Tesla K40m GPUs.

### B. Model training

The validation loss as a function of the wall clock time and epoch number is given in Fig. 3; the best acceleration with minimum validation loss was achieved for $Q = 3$.

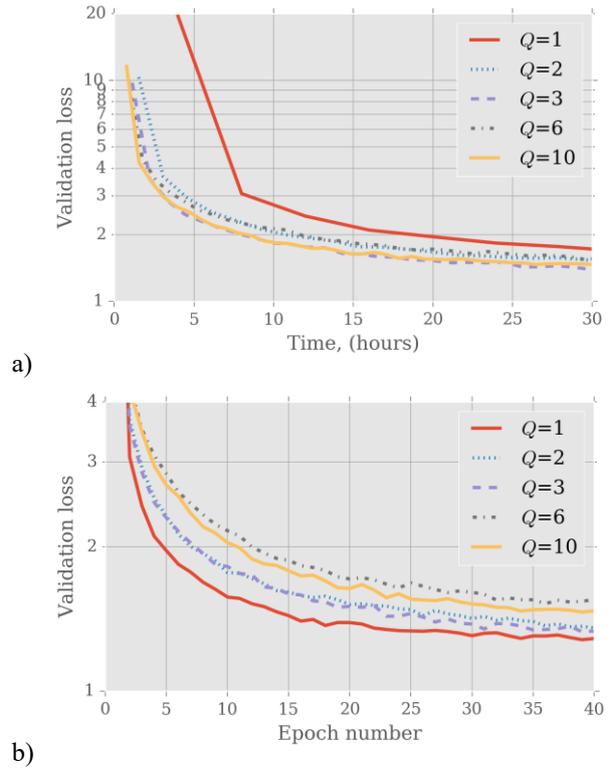

Figure 3. Validation loss of the LSTM model training for different bucket number $Q$. The baseline case corresponds to $Q = 1$

In terms of the wall clock time, the system without bucketing (purely random split of sequences on mini-batches) has the longest epoch time (4 hours per epoch, Fig. 4). The epoch time reduces as $Q$ increases. For the value of $Q = 3$, the speed up factor is close to 4.

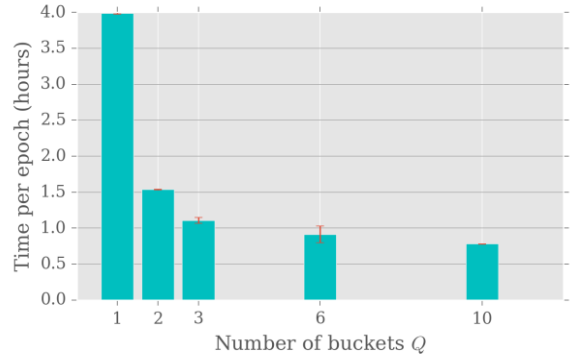

Figure 4. Mean epoch training time of the LSTM model as a function of $Q$ for parallel training on 6 GPUs

From the comparison, we can observe the influence of the sequence buckets on the training speed and loss.

We observed faster convergence of the validation loss especially at the beginning of the training.

The validation loss as a function of wall clock time for different number of used GPUs is presented in Fig. 5a, and as a function of epoch number in Fig. 5b.

The experiment shows that the validation error for 30 hours of training is better by 23% for 6 GPU case (see Fig. 5a, 1 GPU vs 6 GPU comparison).

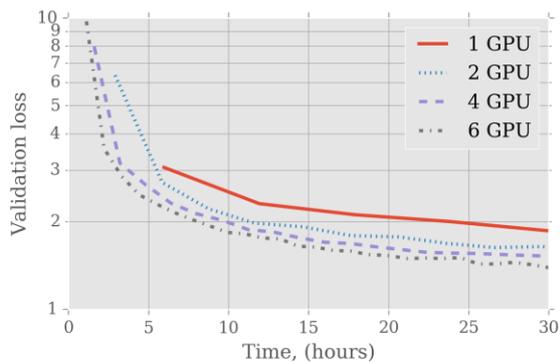

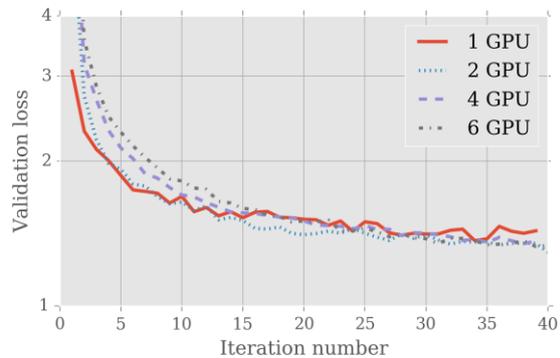

Figure 5. Validation loss of LSTM model training for multiple-GPU architecture. The bucket number $Q = 6$

CONCLUSION

In this paper, the algorithm of RNN training acceleration based on sequence bucketing and multi-GPU data parallelization was presented.

Previous work demonstrated approaches for accelerating RNN gradient descent training involving heuristics for tuning of learning parameters, different network architectures and parallelization techniques. Those solutions, however, did not take into account that the data for perceptual machine learning tasks usually contain input sequences of different length. At the same time, the computational complexity of the training is defined by the longest sequence in the data batch.

The proposed approach improves training speed by clustering sequences into buckets by their length, thus finding a compromise between data structuring and shuffling.

An example of application to online handwriting recognition task with LSTM RNN is given. We obtained the acceleration factor 4 for the number of buckets $Q = 3$. Due to data parallelization in its turn, we observed the reduction of the validation loss by 23% for the same wall clock time compared to the single GPU case. In future work, we plan to investigate different strategies of bucket ordering during training on the model generalization.

This approach may also be useful for LSTM and GRU training in speech recognition, language modelling and other perceptual machine learning tasks.